# Detection of pose orientation across single and multiple axes in case of 3D face images

Parama Bagchi, Debotosh Bhattacharjee, Life Member, IEEE, Mita Nasipuri, Life Member, IEEE and Dipak Kumar Basu, Life Member, IEEE

**Abstract**— In this paper, we propose a new approach that takes as input a 3D face image across X, Y and Z axes as well as both Y and X axes and gives output as its pose i.e. it tells whether the face is oriented with respect the X, Y or Z axes or is it oriented across multiple axes with angles of rotation up to 42º. All the experiments have been performed on the FRAV3D, GAVADB and Bosphorus database which has two figures of each individual across multiple axes. After applying the proposed algorithm to the 3D facial surface from FRAV3D on 848 3D faces, 566 3D faces were correctly recognized for pose thus giving 67% of correct identification rate. We had experimented on 420 images from the GAVADB database, and only 336 images were detected for correct pose identification rate i.e. 80% and from Bosphorus database on 560 images only 448 images were detected for correct pose identification i.e. 80%.abstract goes here.

—————————— ✦ ——————————

## 1 INTRODUCTION

3D face registration has been a crucial and an important field of study in 3D face recognition. In 3D face recognition, registration is a key pre-processing step. The process of aligning facial shapes to a common coordinate system is called registration. Registration is actually the step in 3D modelling to find correspondences between 3D models. The main purpose is to derive a rigid transformation (namely the rotation matrix and the translation vector in the case of rigid objects) that will align the views in a common coordinate system. Registration approaches use the distance between the aligned facial surfaces as a measure of how well faces match between themselves. To align the 3D facial shapes, nearly all state of the art 3D face recognition methods minimize the distance between two face shapes or between a face shape and an average face model. Normally, registration is done on the basis of some landmarks or by aligning to some intrinsic coordinate system defined on the basis of those landmarks. Much of the 3D based face registration and recognition systems presented in the literature uses low noise 3D data in a frontal pose and normalization techniques that do not have good performances in case of different pose variations. In contrast, our method was able to identify and analyze the pose orientation of any 3D face image across all varied poses and also across multiple axes thus enabling an effective way of face registration. In this paper, we have worked on detecting pose alignment for single as well as multi-pose detections and we have performed our experiments on three databases namely the FRAV3D face database, Bosphorus face database as well as GAVADB database. The paper has been organized as follows: - In Section II, some related works have been described. In Section III, a description of our method has been described. A comparison of our method over previous methods has been discussed in Section IV. Experimental results are discussed in Section V and finally in the Section VI, the conclusions are enlisted. In the following Section II, we discuss some related works on 3D face registration and we describe how our method outperforms the already existing method.

## 2 RELATED WORKS

In [1], an analysis of pose variation has been presented but it has not been mentioned about how the estimation of pose orientations with respect to X, Y and Z axes have been estimated and neither any discussions concerning the registration across multiple axes have been

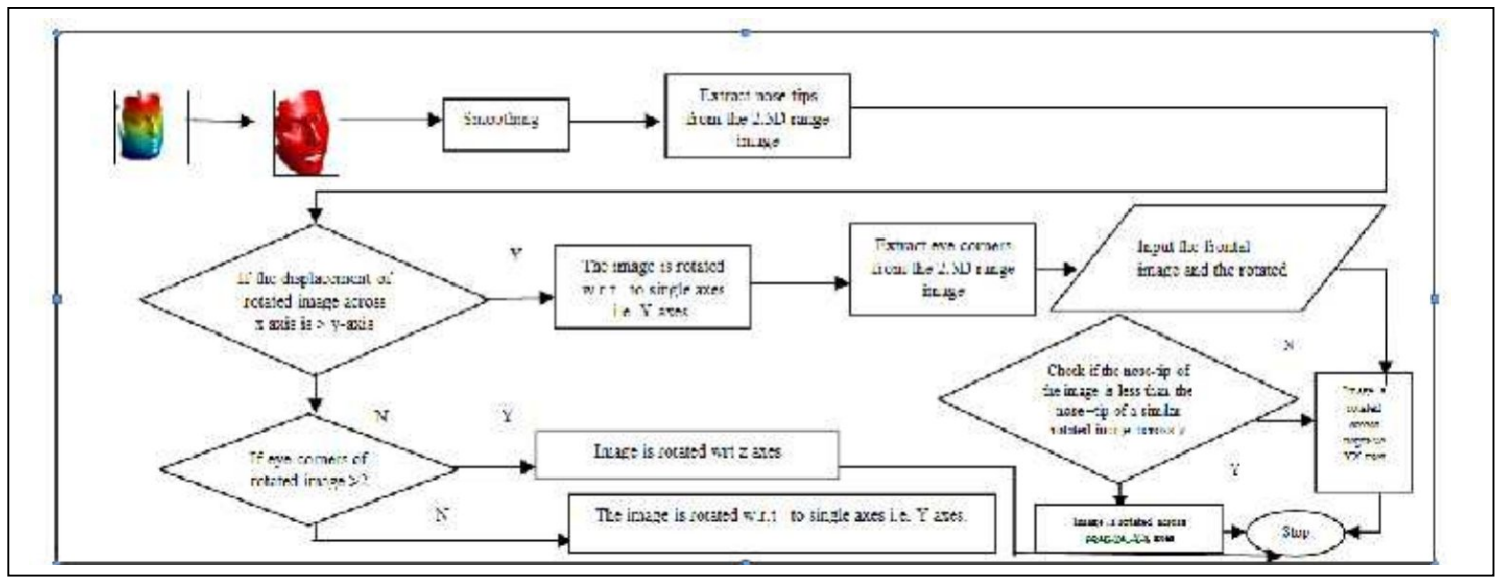
Fig1 Flowchart of the proposed Method

discussed. In [2], the facial symmetry is dealt for determining the pose of 3D faces where, the authors have proposed an effective method to estimate the symmetry plane of the facial surface. But no discussions on how to find the various pose orientations across single as well as multiple axes have been emphasized. In [3] also, a robust approach on 3D registration has been done but no specific discussions about the orientation of the 3D face with respect to the X, Y and Z axes or multiple axes have been highlighted. In [4], the authors have used profile based symmetry assuming an image to be already aligned with respect to X, Y and Z axes but no discussions of how to retrieve the pose alignment with respect to single as well as composite axes have been highlighted. In [5], a discussion of pose has been given with respect to X,Y and Z axes but more specifically the Z corrected poses have been specifically discussed with no emphasis on how to detect orientations across X and Y automatically. The authors discussed anything about how to detect poses across multiple axes. Some other pieces of work include the ones done by Maurer [6] on 3D recognition by extracting 3D texture. Another work was done by Mian [7] on 3D spherical face representation (SFR) used in conjunction with the scale-invariant feature transform (SIFT) descriptor to form a rejection classifier, which quickly eliminates a large number of candidate faces at an early stage for recognition in case of large galleries. This approach automatically segments the eyes, forehead, and the nose regions, which are relatively less sensitive to expressions and matches them separately using a modified iterative closest point (ICP) algorithm.

Another work was done by Queirolo [8] which presented a novel automatic framework to perform 3D face recognition. The proposed method uses a Simulated Annealing-based approach (SA) for range image and registration with the Surface Interpenetration Measure (SIM) in order to match two face images. Another work done by Faltemier [9] is based on 3D face recognition with no discussion on pose variances. In [23], pose was estimated across yaw, pitch and roll but images were captured from several viewpoints at the same time with cameras at known positions. In [24], the authors already assumed that the 3D pose could be approximated by an ellipse and that the aspect ratio of the ellipse was given. In [25], only computation of head orientation and gaze detection was done. In[26], only object identification was done. In all these works, no emphasis about how to find out the pose alignment of any 3D face being input to the system is discussed. In contrast to all the above methods, our pose-orientation method could predict exactly in what pose the 3D face is oriented i.e. whether it is oriented across X, Y or Z axes as well as if there is an orientation across multiple or composite axes, then also it could be deduced as to across which orientation the 3D face is rotated. This is vn.ery important because if the orientation cannot be found out properly then proper registration could not be performed. There are many databases which could provide the poses across X, Y and Z from which we have performed our experiments on FRAV3D, Bosphorus, and GAVADB databases. But, unfortunately there are not many databases which provide 3D face models which are oriented across multiple axes. In this particular paper, FRAV3D as well as GAVADB databases were used for poses across X,Y and Z axes and for X,Y axes respectively but, Bosphorus database was used for poses across multiple axes i.e. for both Y and X axes based composite rotations as well as for poses across X,Y axes.

After applying our algorithm, we are getting comparably good results, in contrast to our previous work stated in [10], and on 560 images only 448 images is giving correct pose identification i.e. identification rate was 80%.

In this paper we have used a geometrical model for determining the pose orientation of a 3D face which would be described in Section III. Fig. 1 shows a brief description of our model for pose detection in the form of a flowchart for finding the pose of a given 2.5D range image. A 2.5D range image as shown as the first picture in Fig-1, is a simplified three dimensional(x,y,z) surface representation that contains at most one depth(z) value. It contains 3D structural information of the surface and by doing so, the illumination problem can be solved. A 2.5D image is actually a good alternative to a 3D image. We hereby want to emphasize the fact that we are considering the images across X, Y, and Z and especially in addition to that a composite axes namely rotations across both Y and X axes for our purposes. The flowchart essentially summarises the entire model of our proposed method in this paper. The proposed model has been discussed in Section III.

## 3 OUR PROPOSED SYSTEM MODEL

The present system for pose detection of 3D range images consists of five steps:-
 i. Pre-processing
ii. Nose-tip estimation by maximum intensity technique
iii. Eye-corners detection by curvature analysis
iv. Alignment of Models
 v. Registration

We prefer to say that the fifth step that is registration will not be taken up in this section of the paper and now we will only be concentrating on pose alignment.

i) Pre-processing:- In this step, every 2.5D range image is subjected to cropping and smoothing. Every 2.5D range image from FRAV3D, Bosphorus or GavaDB database are cropped out by fitting an ellipse to the 2.5 range image. After cropping the range images from o database looked like as in Fig 2. Fig 2 shows so samples across various pose variations that we ha taken for testing for pose detection.

In the next step of pre-processing, all the range images are smoothed by Gaussian filter[11] and outlier is detected by RANSAC algorithm. After smoothing the 3D images obtained with respect to the basic 4 orientations i.e. frontal pose, rotation about X, Y and Z axes are as shown in Fig 4 corresponding to the 2.5D range images as in Fig-3.

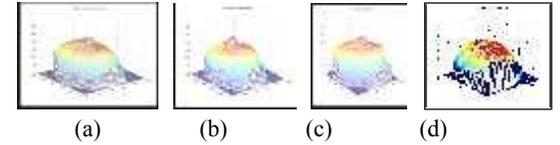

*Fig4. 3D images in frontal pose (a) rotated about X-axis(b),Y-axis(c,)Z axis(d), and YX axis(e) after smoothing.*

ii) Nose-tip estimation using maximum intensity technique [15]:- For the nose tip localization, the method we have used is the maximum intensity concept as the tool for the selection process. Each of the faces (including rotation in any direction in 3D space namely about x-axis, y-axis and z-axis) from the FRAV3D database has been next inspected for localizing the nose tip. A set of points are extracted from both frontal and various poses of face images using a maximum intensity tracing algorithm and the point having the maximum depth intensity was localized to be the nose-tip. The region of interest i.e. the nose tip has been marked on the surface so generated. Fig 5 shows the 3D faces in various poses mentioned, with nose tip localized. The algorithm for localization of nose-tip is given below:-

Algorithm1 : Find_Nose-tip
Input: 2.5D Range Image
Output: - Nose-tip coordinates

```
1:  Initialize max to  0
2:  for I =1 to   width (Image)
3:  for J=1 to   height(Image)
4:  Set val to sum (image (I-1: I+1, J-1: J+1))
5:  Check if val is greater than max.
6:       Set val to val2
7:  end if
8:  end for
9:  end for
```

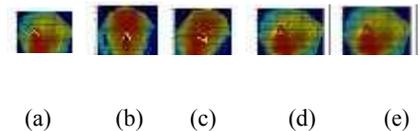

*Fig. 5 : Images showing the nose-tip localized in range images in frontal pose(a), rotated about X-axis(b),Y-axis(c) and Z axis(d) and YX axis(e) after smoothing.*

scheme of nose tip detection works only on poses up to 42º.

iii) Eye corners detection by curvature analysis:- The present face detection method allows the faces, if any, to be freely oriented with respect to the camera plane, the only restriction being that no self-occlusions hide the eyes, the nose, or the area between them. The method analyzes a three-dimensional shot of a real scene given as a range image, that is an image where for each location (i, j) the coordinates are ($x_{ij}$, $y_{ij}$, $z_{ij}$). Curvature of a surface is given by the values of first and second derivatives.

Let S be a surface defined by differentiable real valued function f :-

$$S = \{(x, y, f(x, y))\} \quad \ldots (1)$$

The mean (H) and the Gaussian (K) curvature can be computed, at point (x, y, f(x, y)) can be calculated as follows [10]:-

$$H(x,y) = \frac{(1+f_y^2)f_{xx} - 2f_x f_y f_{xy} + (1+f_x^2)f_{yy}}{2(1+f_x^2+f_y^2)^{3/2}} \quad \ldots (2)$$

$$K(x,y) = \frac{f_{xx}f_{yy} - f_{xy}^2}{(1+f_x^2+f_y^2)^{3/2}} \quad \ldots (3)$$

where fx, fy, fxx, fxy, fyy are the partial derivatives at (x, y). To estimate the partial derivatives, we consider a least squares biquadrate approximation for each element represented by the coordinates(i, j) in the range image:-

$$g_{ij}(x, y) = a_{ij} + b_{ij}(x - x_i) + c_{ij}(y - y_j) + d_{ij}(x - x_i)(y - y_j) + e_{ij}(x - x_i)^2 + f_{ij}(y - y_j)^2 \quad \ldots (4)$$

where the coefficients $a_{ij}$, $b_{ij}$, $c_{ij}$, $d_{ij}$, $e_{ij}$, $f_{ij}$ are obtained by the least squares fitting of the points in a neighbourhood of ($x_i$, $y_j$). Here $g_{ij}$ is the bi-quadratic approximation of the surface and $f_{ij}$ is the partial derivative of the surface. The coefficients of f in ($x_i$, $y_j$) are then estimated by the derivatives of $g_{ij}$. $fx(x_i,y_j) = b_{ij}$, $fy(x_i, y_j) = c_{ij}$, $fxy(x_i, y_j) = d_{ij}$, $fxx(x_i,y_j) = 2e_{ij}$, $fyy(x_i, y_j) = 2f_{ij}$. The mean(H) and the Gaussian curvature(K) can then be estimated as follows:-

$$H(x_{ij}, y_{ij}) = \frac{2(1+c_{ij}^2)e_{ij} - b_{ij}c_{ij}d_{ij} + (1+b_{ij}^2)f_{ij}}{(1+b_{ij}+c_{ij})^{3/2}} \quad \ldots (5)$$

$$K(x_{ij}, y_{ij}) = \frac{4e_{ij}f_{ij} - d_{ij}^2}{(1+b_{ij}^2+c_{ij}^2)^2} \quad \ldots (6)$$

Analyzing the signs of the mean and the Gaussian curvature, we perform what is called an HK classification of the points of the region to obtain a concise description of the local behaviour of the region. Table 1 shows the local behaviour of a 3D surface on the basis of H-K classification table.

**TABLE I**

HK- CLASSIFICATION TABLE.

| Values of H | Values of K | | |
|---|---|---|---|
| | K>0 | K=0 | K<0 |
| H<0 | Elliptical convex | Cylindrical Convex | Hyperbolic Convex |
| H=0 | Impossible | Planar | Hyperbolic Symmetric |
| H>0 | Elliptical concave | Cylindrical Concave | Hyperbolic Concave |

Fig 6 shows the areas of interest i.e. the eye corners taken for our purpose.

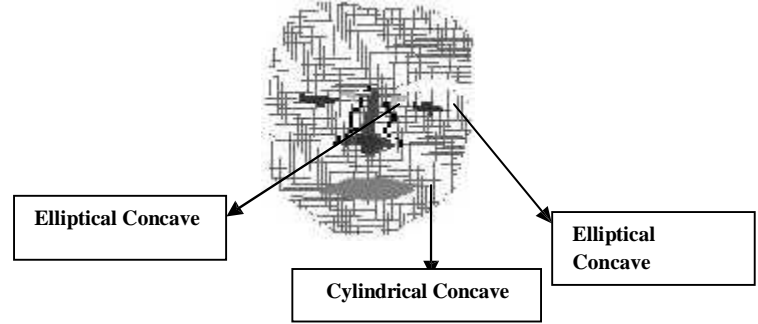

Figure 6. Segmented regions of a face from which facial features are extracted with the curvature map of region.

In this section, the eye-corners have been generated using curvature analysis [12, 13]. In Fig 7, we show the detected eye-corners marked in blue, in frontal pose using curvature analysis.

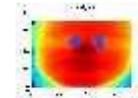

Fig 7. Figure showing eye-corners detected in frontal pose using curvature analysis.

After running the algorithm the output obtained is as shown:-

```
Row   Col   Curvature-value
 29    51    0.000410
 49    50    0.000225
```

The points of highest curvature values are the inner corners of the eye-region. Similarly, the detected eye-corners marked in blue have been shown in Fig 8 for a 3D image rotated about Y-axis.

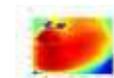

Fig 8. A sample figure showing detected eye-corners for an image in rotated pose rotated about Y axis using curvature analysis.

```
Row Col Curvature-value
 20   53   0.000998
  8   53   0.000336
```

Similarly, the detected eye-corners marked in brown have been shown in Fig 9 for a 3D image rotated about X-axis.

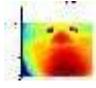

*Fig 9. A sample figure showing detected eye-corners for an image in rotated poses rotated about X axis using curvature analysis.*

| Row | Col | Curvature-value |
|---|---|---|
| 29 | 50 | 0.092934 |
| 51 | 51 | 0.00113 |

Similarly, the detected eye-corners marked in blue, have been shown in Fig 10, for a 3D image rotated about Z-axis.

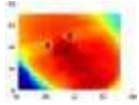

*Fig 10. A sample figure showing detected eye-corners for an image in rotated poses rotated about Z axis using curvature analysis.*

| Row | Col | Curvature-value |
|---|---|---|
| 37 | 51 | 0.000357 |
| 18 | 43 | 0.000184 |

Similarly, the detected eye-corners marked in blue, have been shown in Fig 11, for a 3D image rotated about YX-axis.

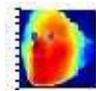

*Fig 11 A sample figure showing detected eye-corners for an image in rotated poses rotated about YX axis using curvature analysis.*

| Row | Col | Curvature-value |
|---|---|---|
| 39 | 65 | 0.002931 |
| 19 | 65 | 0.001577 |

The algorithm that has been used to extract the eye-corners is as given in Algorithm-2.

---

Algorithm 2: Detect_ EyeCorners
Input   : 3D Image rotated across X, Y, Z and YX axes
Output : Eye-corners returned.

---

1:   Generate H and K curvature map.
2:   Initialize the value of p to 0
3:   for I=1 to width of Image
4:   for J=1 to height of Image
5:   if  H >0 and K>0.0001
6:       arr[p]=K
7:        p=p+1
8:   end if
9:   end for
10: end for
11: Sort the array in descending order.
12: Extract the first and second elements of the array named arr as the final Gaussian curvature values.

---

For detecting the nose-tip, Gaussian curvature map was first thresholded for values above 0.0001.

iv) Alignment of Models:- Here, in the last and final section the alignment techniques adopted by us to infer the axes across which the 3D face is rotated i.e. single or multiple axes has been discussed.

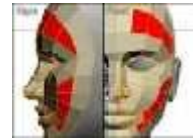

*Fig 12. A triangulated face image. [14]*

From the Figs 8, 9 and 10, it can be inferred that:-

- If a 3D face is rotated with respect to X axis, both the eyes must lie on the same horizontal line. Deviation errors are possible but definitely within a minimum threshold.
- If a 3D face is rotated with respect to Y axis, both the eyes must lie on the same horizontal line. Deviation errors are possible but definitely within a minimum threshold.
- If a 3D face is rotated with respect to Z axis, both the eyes must not lie on the same horizontal line. So as stated above, we have tried to segregate the X axis and Y axis from Z axis.

Now, we propose a method to segregate the X and Y axes:-

- If a 3D face is rotated with respect to X axis, deviation of nose-tips with respect to Y axes will be more than X axes.
- If a 3D face is rotated with respect to Y axis, then deviation of nose-tips with respect to X axes will be more than Y axes.

Now, let us propose a method to solve composite pose across more than 2 axes. The anomaly of this method is that there are very few databases which actually provide pose-variations across multiple axes thus making it difficult to detect some special cases of multi-pose variations correctly. Since we have mentioned earlier that we have basically worked on three different databases, FRAV3D, GavaDB and Bosphorus databases, we must specify that the only database where we could actually find some interesting multiple pose variations is the Bosphorus database[19]. Bosphorus database provides a very small amount of multiple pose variations i.e. only two poses variations across composite Y axes and X axes per individual. So, basically we have 120 multiple pose variations to design the system. Now, let us consider the following images as shown in Fig 13. The images are basically 2D images corresponding to the range images as shown in Fig 3(b).

- Now, the position of the eye-lids would have clearly stated the location of a multiple pose variation i.e. if a face is rotated with respect to positive combined Y axes and X axes, the position of eye-lids would be oriented upwards as shown in the figure 13(a).

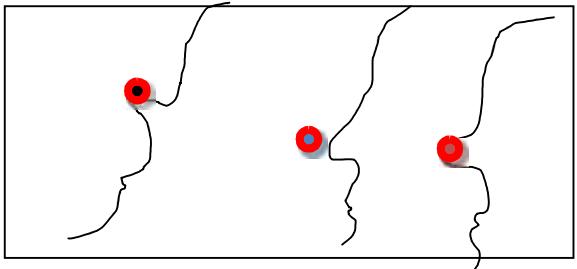

Fig 14. A plot showing the profile curves corresponding to the 2D images of Fig 13.

We have plotted the nose-tip in case of the three profile curves corresponding to the Fig 13. Now as can be seen

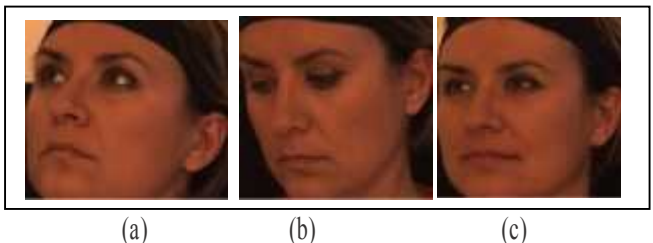

(a)      (b)      (c)

Fig 13. 2D images rotated across positive YX axes (a) negative YX axes (b) and single y axes(c).

If a face is rotated with respect to positive YX axes the position of eye-lids would be oriented upwards as shown in the figure 13(b). But, in case of 3D face, since eye-lids have a very small or near to zero depth intensity, it is very difficult to detect the orientations as well as the curvature-values of the eye-lids. If the gradient of the eye-lids surfaces are taken, two major problems will arise:- from Fig 13 the alignment with respect to the X axis is maximum of 10º. So the nose-tip will not much elevated upwards or oriented downwards. The idea is, we have to input a image rotated across Y axes and then compare how much displacement has the resulting multiple pose variated image has with respect to that rotated image across Y axes. In Algorithm-1, we have already detected the nose-tip of a 2.5D range image, so we apply the same algorithm to extract to nose-tip across multiple pose variations. As can be very well seen in Fig 14 an image rotated with respect to positive combined Y axes and X axes has its nose tip more elevated than an image rotated with respect to only Y axes. Again an image rotated with respect to positive combined Y and X axes has its nose tip less elevated than an image rotated with respect to only Y axes. At the end we have to distinguish between an image rotated by single axis and multiple axes by the displacement of the nose-tip only. We are summarising this idea in Algorithm 3 for pose detection.

Algorithm 3:- Pose_Alignment_XYZ
Input: - 3D face Image in rotated pose and a 3D face image in unknown multiple pose.
Output: - Pose Alignment returned with respect to X, Y, and Z axes

1) It is very difficult to capture the minute variations of the eye-lids of an individual
2) Even if the eye-lids are captured, gradient not be captured or calculated to the nearest perfection.

So, it is better to come up with alternate solutions. Let us look at Fig 14 showing the facial curves of the profile corresponding to Fig 13.

1: From the databases choose any frontal image of a person.
2: Find out the nose-tip of the person in the image by maximum intensity technique.
3: Find out the eye-corners by curvature analysis.
4: Find out the nose-tip of the person with the rotated pose by maximum intensity technique
5: Find out the eye-corners of the person with the rotated pose i.e. input image by curvature analysis [13,14]. Assign the X and Y coordinates of the eye-corners of the rotated 3D face image which is in an unknown pose variation to variables x1coordf and x2coodrf.
7: Also, assign the xcoordinates of the nose-tips of frontal
and the unknown pose to variables x1 and x2
8: Also, assign the ycoordinates of
the nose-tips of frontal and multiple
pose pose to variables y1 and y2
9: Assign diff = x1coordf - x2coordf
10: if diff >e then
11: Output that the 3D face is rotated with respect to Z axis
12: else
13: if x1 >x2 or if x1 < x2
14: if abs(x1 - x2) > = abs (y1 – y2 )
15: Face is rotated wr.t. Y axis
16: Call multipose detection(Roatated 3D image)
17: end if
18: end if
19: if $y_1 > y_2$ or $y_1 < y_2$
20: if abs $(y_1 – y_2) >= $ abs$(x_1 - x_2)$
21: Print 3D face rotated with respect to X axis
22: end if
23: end if
24: end if

Here, we note that, the value of the parameter e in our algorithm is set to two, because if we notice the Figures 7, 8, 9, 10 the eye-corners in case of X and Y axes vary at the most by a degree of two. Hence, the value of e has been set to two because eyes are normally at the same level. Now, Algorithm 3 segregates the three principal axes in case of 3D images i.e. X Y and Z. Algorithm 4 i.e. Multi_pose_detection function tries to find out a way so as to detect multi-pose alignment i.e. alignment with respect to more than a single axes, in our case it would be the YX axes.

---

Algorithm_4:-Multi_pose_detection
Require: A 3D face image in an unknown multiple pose orientation

---

1: Detect the nose-tip of the input 3D image rotated w.r.t. Y axes by Algorithm 1.
2 : Detect the nose-tip of the unknown multiple pose 3D face image by Algorithm 1.
3: y_roty = y-axis of rotated 3D image across Y axis
4: x_rotx= x-axis of rotated 3D image across X-axis.
5: y_rotmultiple pose = y-axis of rotated multiple pose 3D image
6: x_rmultiple pose = x-axis of rotated multiple pose 3D image
7: if y_rotmultiple pose > y_roty
8: if abs(y_rotmultiple pose-y_roty) > 3
9: 3D face image is rotated w.r.t. positive YX axes
10: else
11: 3D face image is rotated w.r.t. negative YX
12: End

With this, we come to the end of the pose detection algorithm, and also there are certain assumptions regarding this algorithm which has been enlisted in the following section. Also, all the experiments have been performed on the bosphorus database because FRAV3D and GAVADB does not have a single image rotated about YX axes.

Here, it is assumed that the nose-tips of any rotated image across Y and its corresponding rotated image across YX axes must be at least at a difference of a factor of three for our algorithm to work correctly. As depicted by our experimental results this is working correctly on Bosphorus database and giving us detection rate of 80%. Though it must be argued that only detecting the nose-tip might not be a very robust method in some cases, but still it has actually been proved much useful. Henceforth, in the section 4, we present a comparative analysis of our present method with our previous technique [9].

## 4 COMPARATIVE ANALYSIS AND EXPERIMENTS WITH DIFFERENT POSE ALIGNMENT TECHNIQUES AND OUTLIERS DETECTION

Here, we shall present a comparative analysis of our technique over our previous work mentioned in [10] and another very significant work done in [16] by the authors. The table 2 describes extended comparison of our present technique and the advantage over [10] and [16]. From the table it is obvious that our method of pose analysis across single as well as multiple axes performs much better than the two other methods [10, 16] discussed above. We have already discussed our method of pose detection. In this section we have specified in the table 2 itself the method used by the authors in [16] and we have compared how our method excels in comparison to the technique described in [16].

# TABLE II

Performance evaluation of present proposed method over other alignment

| Parameter Measures | Pose detection across single as well as multiple axes | Pose detection across single axis [10] | SFR Model [16] |
|---|---|---|---|
| Outlier Detection Technique | Gaussian Filter and RANSAC | Gaussian filter only used for smoothing and no outlier detection used | Cropping of face and outlier detection by RANSAC algorithm |
| Complexity Analysis | $O(n^2)$ | $O(n^2)$ | Not specified |
| Databases used | Bosphorus, GAVADB, FRAV3D | Only FRAV3D | Only FRGV v2.0 |
| Types of poses detected | Poses detected across X, Y and Z axes and YX axes | Poses detected X,Y and Z axes | Method works correctly for only near frontal images |
| Angles of elevations | Up to 42° only from all the databases | Up to 40° only | Up to 6-7° only |
| Performance evaluation (in terms of Registration/recognition/detection) | 1. FRAV3D:- On 848 3D faces, 566 3D faces were correctly recognized for pose thus giving 67% of correct identification rate. 2. GAVADB:- On 420 images only 336 is giving correct pose identification rate i.e. 80%. 3. Bosphorus:- On 560 images only 448 images is giving correct pose identification i.e. 80% | 1. FRAV3D:- On 848 3D faces, 566 3D faces were correctly recognized for pose thus giving 67% of correct identification rate. | 1. FRGC Ver. 2.0:-dataset (9,500 2D/3D faces) show that our algorithm outperforms 99.47% and 94.09% identification rates for SFR. 2. Identification rates of 98.03% and 89.25% for probes with neutral and non-neutral expression were obtained for SFR. |

## 5 EXPERIMENTAL RESULTS

In this section, we present an analysis of how does our proposed method for alignment analysis work on the FRAV3D[17] database. From tables III to table VII, we have analysed how our method for pose detection works on FRAV3D, GavaDB and the Bosphorus database. Fig-15 shows snapshots of 2D images, taken from the FRAV3D database rotated about X-axis.

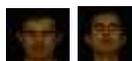

*Fig 15. Images rotated about X axis from the FRAV3D Database.*
The fig 16 shows snapshots of 2D images, taken from the FRAV3D database rotated about Y-axis.

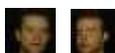

*Fig 16. Images rotated about Y axis from the FRAV3D Database.*
The fig 17 shows snapshots of 2D images taken from the FRAV3D database for images rotated about Z-axis.

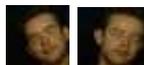

*Fig 17. Images rotated about Z axis from the FRAV3D Database.*
The experimental results are enlisted in the Table III.

## TABLE III
Detection of Pose Alignment in various Angles in FRAV3D Database

| Sl.no | A | B | C | D | E | F | G | H | I |
|---|---|---|---|---|---|---|---|---|---|
| 1 | +5 | 70 | 48 | +10 | 35 | 23 | +18 | 35 | 22 |
| 2 | -5 | 70 | 50 | -10 | 35 | 23 | -18 | 35 | 22 |
| 3 | +18 | 70 | 50 | +38 | 35 | 23 | +38 | 35 | 23 |
| 4 | -18 | 70 | 48 | -38 | 35 | 23 | -38 | 35 | 24 |
| 5 | +40 | 72 | 47 | +40 | 36 | 25 | +40 | 36 | 24 |
| 6 | -40 | 72 | 46 | -40 | 36 | 25 | -40 | 36 | 22 |

Next, we present our results for pose detection for GAVADB[18] database. It is to specify that neither GAVADB nor FRAV3D database has any pose variations across XY, YX or YZ axes. So, we had to perform the experiment on single axes only. The fig 18 shows the figures from GavaDB database oriented across Y axes.

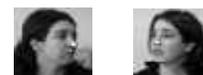

(a)    (b)

Figure 18. Face-images rotated about Y axis from GAVADB Database.

The fig 19 shows the figures from GAVADB database oriented across X axes.

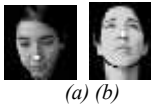
(a) (b)

Figure 19. Face-images rotated about X axis from GAVADB Database.

| Sl.no | A | B | C | D | E | F |
|---|---|---|---|---|---|---|
| 1 | +18 | 105 | 84 | +38 | 105 | 84 |
| 2 | -18 | 105 | 84 | -38 | 105 | 84 |

Here the notations A,B,C,D, E,F,G,H, I stands for:-
A:- Angles of Alignment with respect to X axis in the GAVADB database
B:- Number of 3D images aligned against X-axes in the GAVADB database.
C:- Number of 3D alignments detected correctly by the present algorithm in the GAVADB database.
D:- Angles of Alignment with respect to Y axis in the GAVADB Database
E:- Number of 3D images aligned against Y-axes in the GAVADB database.
F:- Number of 3D alignments detected correctly by the present algorithm in the GAVADB database.

Next, results for pose detection for Bosphorus [20] database are presented below. It is especially important to mention here that we have figures rotated around composite Y axes and X axes in the Bosphorus database. There are two composite rotations per individual in the Bosphorus database.

Here the notations A,B,C,D, E,F,G,H, I stands for:-
A:- Angles of Alignment with respect to X axis in the FRAV3D database
B:- Number of 3D images aligned with respect to X-axes in the FRAV3D database
C:- Number of alignments detected correctly by the present algo in the FRAV3D database
D:-Angles of Alignment with respect to Y axis in the FRAV3D database
E:- Number of 3D images aligned with respect to Y-axes in the FRAV3D database
F:- Number of alignments detected correctly by the present algo in the FRAV3D database
G:-Angles of Alignment with respect to Z axis in the FRAV3D database
H:- Number of 3D images aligned with respect to Z-axes in the FRAV3D database
I:- Number of alignments detected correctly by present algorithm in the FRAV3D database

Next, results for pose detection for Bosphorus [20] database are presented below. It is especially important to mention here that we have figures rotated around composite Y axes and X axes in the Bosphorus database. There are two composite rotations per individual in the Bosphorus database.The Fig-20 shows samples taken from the Bosphorus database for images rotated about X-axis.

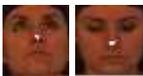

Figure 20. Face-images rotated about X axis from Bosphorus Database.

The Fig-21, shows samples taken from the Bosphorus database for images rotated about Y-axis.

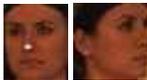

Figure 21. Face-images rotated about y axis from Bosphorus Database

The Fig-22 shows some samples taken from the Bosphorus database for images rotated about YX composite axis.

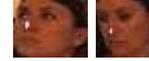

Figure 22. Face-images rotated about YX axis from Bosphorus Database

The experimental results for pose alignment detection with respect to X axes are enlisted in the Table V.

**TABLE V**
Detection of Pose Alignment in various Angles across X axis

| Sl.no | A | B | C |
|---|---|---|---|
| 1 | +5 | 80 | 70 |
| 2 | -5 | 80 | 70 |

Here the notations A, B, C stands for:-
A:- Angles of Alignment with respect to X axis in the Bosphorus database

B:- Number of 3D images aligned with respect to X-axes in the Bosphorus database

C:- Number of alignments detected correctly by the present algorithm in the Bosphorus database

**TABLE VI**
Detection of Pose Alignment in various Angles across Y axis

| Sl.no | D | E | F |
|---|---|---|---|
| 1 | +10 | 80 | 60 |
| 2 | +20 | 80 | 80 |
| 3 | +30 | 80 | 70 |

D:- Angles of Alignment with respect to Y axis in the Bosphorus database
E:- Number of 3D images aligned with respect to Y-axes in the Bosphorus database
F:- Number of alignments detected correctly by the present algorithm in the Bosphorus database

**TABLE VII**
Detection of Pose Alignment in various Angles across yx axis

| Sl.no | G | H | I |
|---|---|---|---|
| 1 | +42º across y axis, then 10º across positive x axis | 80 | 50 |
| 2 | +42º across y axis, then -10º across positive x axis | 80 | 48 |

G:- Angles of Alignment with respect to composite Y and X axes in the Bosphorus database
H:- Number of 3D images aligned with respect to composite Y and X axes in the Bosphorus database
I:- Number of alignments detected correctly by the present algorithm in the Bosphorus database

The paper depicts a novelty over our previous work in [10] since we have tried to implement our model for pose alignment over multiple databases and also across composite axes. In literature very little work has been done in this section.

## 6 PERFORMANCE OF THE PRESENT METHOD IN TERMS OF QUANTIFICATION

In this section, we shall quantify our results with the standard results obtained in terms of nose-tip detection in terms of other similar works. In Table VIII, we detect the rate of correct classification in terms of our method proposed in this paper, with standard nose-tip localization techniques and we try to find out how our method is superior with respect to the already prevailing techniques.

**TABLE VIII**
**Performance Results**

| Pose-angles | Success Rate of Nose Tip Detection for Non-Frontal Faces |
|---|---|
| Method in [21] for detection of nose-tip | 5% |
| Densest Nose Tip Candidate Method[22] | 15% |
| Our proposed method | Max 80% |

## 7 CONCLUSION AND FUTURE SCOPE OF THE WORK

In this present approach, the previous approach for pose detection has been extended to detect poses across multiple axes. The algorithm gives us sufficiently good results over our previous work. But, at the same time it is true that it is not the best result because possibly of the presence of many outliers. This problem would be addressed in future. As a part of our future work, we would like to address the problem of 3D face registration. Here, in this present work, we have already addressed the problem of small poses up to 42º but we are still to address poses across 90º as a part of our future work. We actually need to mention here that the problem of curvatures that we have addressed here, are still to be implemented across sufficiently large poses

## RERERENCES

## ACKNOWLEDGEMENTS

Authors are thankful to a project entitled "Development of 3D Face Recognition Techniques Based on Range Images," supported by DIT, MCIT, Govt. of India, at Department of Computer Science and Engineering, Jadavpur University, India for providing necessary infrastructure to conduct experiments relating to this work.

## APPENDIX B

**Parama Bagchi** received BTECH(CSE)and M.TECH(Computer Technology) degree from BCET under WBUT and Jadavpur University in 2005 and 2010 respectively. She received the Gold medal from Jadavpur University for securing the highest marks in M.Tech(ComputerTechnology) in 2010. She has worked as a lecturer in CSE in various private engineering colleges. Presently she is posted as an Assistant Professor in MCKV Institute of Engineering, Liluah, Howrah in CSE Department. Her current research interests include Image Processing and Face Recognition. She is currently pursuing her research work in Jadavpur University.

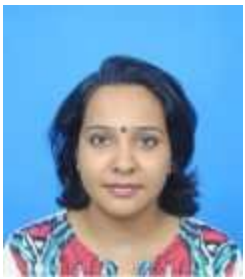

**Debotosh Bhattacharjee** received the MCSE and Ph. D.(Eng.) degrees from Jadavpur University, India, in 1997 and 2004 respectively. He was associated with different institutes in various capacities until March 2007. After that he joined his Alma Mater, Jadavpur University. His research interests pertain to the applications of computational intelligence techniques like Fuzzy logic, Artificial Neural Network, Genetic Algorithm, Rough Set Theory, Cellular Automata etc. in Face Recognition, OCR, and Information Security. He is a life member of Indian Society for Technical Education (ISTE, New Delhi), Indian Unit for Pattern Recognition and Artificial Intelligence (IUPRAI), and senior member of IEEE (USA).

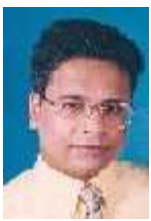

**Mita Nasipuri** received her B.E.Tel.E.,M.E.Tel.E,and Ph.D. (Engg.) degrees from Jadavpur University, in 1979, 1981 and 1990, respectively. Prof. Nasipuri has been a faculty member of J.U since 1987. Her current research interest includes image processing, pattern recognition, and multimedia systems. She is a senior member of the IEEE, U.S.A., Fellow of I.E (India) and W.B.A.S.T, Kolkata, India.

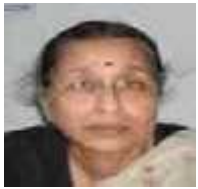

**Dipak Kumar Basu** received his B.E.Tel.E., M.E.Tel., and Ph.D. (Engg.) degrees from Jadavpur University, in 1964, 1966 and 1969 respectively. Prof. Basu has been a faculty member of J.U from 1968 to January 2008. He was an A.I.C.T.E. Emiretus Fellow at the CSE Department of J.U. His current fields of research interest include pattern recognition, image processing, and multimedia systems. He is a senior member of the IEEE, U.S.A., Fellow of I.E. (India) and W.B.A.S.T., Kolkata, India and a former Fellow, Alexander von Humboldt Foundation, Germany.

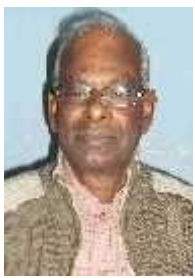